\documentclass{elsarticle}
\usepackage{hyperref}
\usepackage{lineno,hyperref} \modulolinenumbers[5]
\usepackage{graphicx}
\usepackage{mathtools}
\usepackage{units}
\usepackage[ruled,vlined]{algorithm2e}
\usepackage{physics} 

\usepackage{color}

\definecolor{green}{rgb}{0,0.7,0}

\providecommand{\martinc}[1]{}                  

\providecommand{\sub}[1]{_{\mathrm{#1}}}  

\journal{Transportation Research Part B}

\begin{document}
	
	\begin{frontmatter}
		
		\title{Formulation and validation of a car-following model based on deep
			reinforcement learning}
		
		
		\author[firstAddress]{Fabian Hart\corref{corrAuthor}}
		\cortext[corrAuthor]{Corresponding author}
		\ead{fabian.hart@tu-dresden.de}
		\author[firstAddress]{Ostap Okhrin}
		\author[firstAddress]{Martin Treiber}

		\address[firstAddress]{Technische Universität Dresden\\
			01062 Dresden\\
			Germany}

		\begin{abstract}		
			We propose and validate a novel car following model based on deep reinforcement learning. Our model is trained to maximize externally given reward functions for the free and car-following regimes rather than reproducing existing follower trajectories. The parameters of these reward functions such as desired speed, time gap, or accelerations resemble that of traditional models such as the Intelligent Driver Model (IDM) and allow for explicitly implementing different driving styles. Moreover, they partially lift the black-box nature of conventional neural network models. The model is trained on leading speed profiles governed by a truncated  Ornstein-Uhlenbeck process reflecting a realistic leader's kinematics.
			
			This allows for arbitrary driving situations and an infinite supply of training data. For various parameterizations of the reward functions, and for a wide variety of artificial and real leader data, the model turned out to be unconditionally string stable, comfortable, and crash-free. String stability has been tested with a platoon of five followers following an artificial and a real leading trajectory. A cross-comparison with the IDM  calibrated to the goodness-of-fit of the relative gaps showed a higher reward compared to the traditional model and a better goodness-of-fit.
		\end{abstract}

		
	\end{frontmatter}
	
	
	\section{Introduction}
	Autonomous driving technologies offer promising solutions to
	improve road safety, where human errors account for 94\% of the total
	accidents \citep{vehicleCrashSurvey2015}.  
	Moreover, they have the potential to reduce energy consumption and emissions. 
	However, autonomous driving is a challenging task
	since traffic flow can be dynamic and unpredictable.
	On the way to fully autonomous driving, Advanced Driver Assistance Systems
	(ADAS) have been developed to solve tasks like car following, emergency
	braking, lane keeping, or lane changing. 
	Since Deep Learning methods that use deep neural networks have been demonstrated to surpass humans
	in certain domains, they are also adopted in the area of autonomous
	driving.
	Especially Deep Reinforcement Learning (DRL), which combines the power
	using deep neural networks with Reinforcement Learning, has shown its potential in a wide variety of autonomous driving tasks. 
	In \cite{OnRampMerge2018} and \cite{OnRampMerge2020}, DRL is used to
	guide an autonomous vehicle safely via an on-ramp to the freeway. Another approach is to manage traffic of autonomous vehicles at intersections, optimizing safety and efficiency such as in \cite{intersection1}, \cite{intersection3} and \cite{intersection2}.
	In \cite{LangeChange1}, DRL is used to solve lane-changing maneuvers.
	
	A further task in autonomous driving is to model the vehicle behavior
	under car-following scenarios, where suitable accelerations must be
	computed to achieve safe and comfortable driving. Approaches
	for solving this task are classical car-following models, such as the
	Intelligent Driver Model (IDM) \citep{Opus} or stochastic car-following
	models such as that of \cite{Treiber2018stochIDM_TRB}. Furthermore, data-driven approaches used Deep Learning methods to train a car-following model based on experimental car-follower data, such as in \cite{Chong2011SimulationOD}, \cite{ZHOU2017245}, \cite{HumanLikeCF} and \cite{HumanLikeCF2}. The downside of this approach is that the model tries to emulate human driver behavior, which can still be suboptimal.
	
	To overcome this issue, DRL methods train non-human car-following models that can optimize metrics such as safety, efficiency, and comfort. 
	One approach is to train on scenarios where the leading vehicle
	trajectory, used for training, is based on experimental data, such as
	in \cite{SafeEfficientAndComfortable} or
	\cite{HumanLikeAutonomouCF}. Similar approaches suggest a standardized
	driving cycle serving as a leading vehicle trajectory, such
	as \cite{ComparisonRLvsMPC} or \cite{CFelectricVehicle}
	who use the New European Driving Cycle.
	A disadvantage coming along with these approaches is that the
	performance decreases for scenarios that are not in the scope of the training data, indicating inadequate machine learning generalization \citep{ComparisonRLvsMPC}. 
	
	Another issue of car-following models is string stability. There are
	several studies focusing on dampening traffic oscillations by using a
	sequence of trained vehicles, such as these by \cite{qu2020jointly}, \cite{DissipatingStopAndGoWaves}, and \cite{DampenStopAndGoTraffic}. \cite{Patrol} developed a car-following model to solve 'phantom' traffic jams by obtaining information of multiple vehicles ahead. \cite{EndToEndCF} proposed a DRL end-to-end car-following model based on video input.
	
	All the mentioned DRL car-following models have three disadvantages in
	common: First, the acceleration range is limited in a way that
	full-braking maneuvers are not considered. This results in models that are just designed for non-safety-critical scenarios. Second, these models just consider car-following scenarios, while free driving or the transition between both is not reflected in the reward function. Third, the trained models have just been validated on data that is similar to the training data set so that the generalization capabilities cannot be proven. 
	
	This motivated us to design a model that overcomes all of these issues.  Specifically,  we propose a RL-based model with the following features:

	\begin{itemize}
		\item applicable for free-driving and car-following situations with safe, smooth, and  comfortable transitions between these states;
		\item wide range of possible
		accelerations leading to a collision-free behavior
		also in safety-critical situations such as a full-braking maneuver of the leader;
		\item being trained on leading trajectories
		based on an \cite{OU} process with parameters reflecting the kinematics of real
		drivers. This leads to high generalization capabilities and
		model usage in a wider variety of traffic
		scenarios. Moreover, the supply of learning data is
		unlimited;
		\item different driver characteristics to be modeled by adjusting the parameters of the reward function;
		\item string stability even in extreme situations.
	\end{itemize}
	To our knowledge, no car-following model has been proposed that has these features.
	
	Another feature of this work is a thorough validation of the above-mentioned properties in scenarios based on both synthetic and real trajectory data, bringing the model to its limits. 
	In all cases, the model proved to be accident-free and string stable.
	In a further experiment, the proposed model is compared to the IDM calibrated on the same data. The results indicate a better performance of the proposed model.
	
	This work is structured as follows. In Section \ref{sec:RLBackground},
	we introduce some basic RL background as well as our modularized RL
	approach, followed by a detailed description of our two sub-modules:
	the Free-Driving Policy in Section \ref{sec:FreeDrivingPolicy} and the
	Car-Following Policy in Section
	\ref{sec:CarFollowingPolicy}. In Section \ref{sec:validation}, we evaluate our RL
	strategy with respect to safety, \emph{stability}, and comfort aspects
	before we conclude in Section \ref{sec:conclusion}.

	\section{Reinforcement Learning Background}
	\label{sec:RLBackground}
	
	The follower is controlled by a Reinforcement
	Learning (RL) agent. By interaction with an environment, the RL agent
	optimizes a sequential decision-making problem. At each time step
	$t$, the agent observes an environment state $s_t$ and, based on that state, selects
	an action $a_t$. After conducting action $a_t$, the RL agent receives
	a reward $r(a_t,s_t)$. The agent aims to learn an optimal state-action
	mapping policy $\pi$ that maximizes the expected accumulated
	discounted reward
	\begin{equation}
		\label{Rt}
		R_{t}=\sum_{k=0}^{\infty} \gamma^{k} r_{t+k},
	\end{equation}
	where $\gamma = (0,1]$ denotes the discount factor and 
	$\gamma^k r_{t+k}$ the expected contribution $k$ time steps ahead. 
	\subsection{\label{RL-algorithm}RL algorithm}
	In various similar control problems, the Deep Deterministic Policy
	Gradient (DDPG) Algorithm has been used and proven to perform well on
	tasks with a deterministic and continuous action and a
	continuous state space, such as in
	\cite{SafeEfficientAndComfortable}, \cite{ComparisonRLvsMPC} or
	\cite{HumanLikeAutonomouCF}. The original work can be found in
	\cite{DDPG}. DDPG is an actor-critic method, that uses an actor network
	$\mu\left(s \mid \theta^{\mu}\right)$ with the network weights $\theta^{\mu} $
	to output an action $a$ based on a given state $s$ and a critic network
	$Q\left(s, a \mid \theta^{Q}\right) $ with the network weights  $\theta^{Q}$ to
	predict if the action is good or bad, based on a given state $s$ and
	action $a$. 
	To achieve better training stability, DDPG uses target networks  $\mu^{\prime}$ and $Q^{\prime}$ for the actor and critic networks.
	These target networks have the same architecture as the main networks, only differing in the network parameters. While training, these networks are updated by
	letting them slowly track the main networks
	\begin{align}
		\theta^{Q^{\prime}} & \leftarrow \tau \theta^{Q}+(1-\tau) \theta^{Q^{\prime}}, \\
		\theta^{\mu^{\prime}} & \leftarrow \tau \theta^{\mu}+(1-\tau) \theta^{\mu^{\prime}},
	\end{align}
	where $\tau$ defines the soft target update rate.
	This approach greatly enhances the stability of learning.
	Furthermore DDPG uses Experience Replay, that implements a Replay Buffer $R$, where a list of tuples $\left(s_{i}, a_{i}, r_{i}, s_{i+1}\right)$ for $i \le t$ are stored. Instead of learning from the most recent experience, DDPG learns from sampling a mini-batch from the experience buffer. To implement better exploration by the actor network, DDPG uses noisy perturbations, specifically an autoregressive process with Gaussian innovations. The entire DDPG algorithm is shown in Algorithm~\ref{alg:ddpg}. 
	
	\begin{algorithm}
		\label{alg:ddpg}
		\small
		Randomly initialize critic network $Q\left(s, a \mid \theta^{Q}\right) $ and actor network $\mu\left(s \mid \theta^{\mu}\right)$ with weights $\theta^{Q}$ and $\theta^{\mu} $  \\
		Initialize target network $Q^{\prime}$ and $\mu^{\prime}$ with weights $\theta^{Q^{\prime}} \leftarrow \theta^{Q}, \theta^{\mu^{\prime}} \leftarrow \theta^{\mu}$\\
		Initialize Replay Buffer $R$\\
		\For{$episode = 1$ \KwTo $M$}{
			Initialize a random process $\mathcal{N}$ for action exploration\\
			Receive initial observation state $s_{1}$ \\
			\For{$t = 1$ \KwTo $T$}{
				Select action $a_{t}=\mu\left(s_{t} \mid \theta^{\mu}\right)+\mathcal{N}_{t}$ according to the current policy and exploration noise\\
				Execute action $a_{t}$ and observe reward $r_{t}$ and observe new state $s_{t+1}$\\
				Store transition $\left(s_{t}, a_{t}, r_{t}, s_{t+1}\right)$ in $R$\\
				Sample a random minibatch of $N$ transitions $\left(s_{i}, a_{i}, r_{i}, s_{i+1}\right)$ from $R$\\
				Set $y_{i}=r_{i}+\gamma Q^{\prime}\left\{s_{i+1}, \mu^{\prime}\left(s_{i+1} \mid \theta^{\mu^{\prime}}\right) \mid \theta^{Q^{\prime}}\right\}$\\
				Update critic by minimizing the loss:
				$L=\frac{1}{N} \sum_{i}\left\{y_{i}-Q\left(s_{i}, a_{i} \mid \theta^{Q}\right)\right\}^{2}$\\
				Update the actor policy using the sampled policy gradient:
				$$
				\nabla_{\theta^{\mu}} J \approx \frac{1}{N} \sum_{i} \left. \nabla_{a} Q\left(s, a \mid \theta^{Q}\right)\right|_{s=s_{i}, a=\mu\left(s_{i}\right)} \left.\nabla_{\theta^{\mu}} \mu\left(s \mid \theta^{\mu}\right)\right|_{s_{i}}
				$$
				Update the target networks:
				$$
				\begin{aligned}
					\theta^{Q^{\prime}} & \leftarrow \tau \theta^{Q}+(1-\tau) \theta^{Q^{\prime}} \\
					\theta^{\mu^{\prime}} & \leftarrow \tau \theta^{\mu}+(1-\tau) \theta^{\mu^{\prime}}
				\end{aligned}
				$$
			}
		}

		\caption{DDPG algorithm according to \cite{DDPG}}
	\end{algorithm}

	\subsection{\label{MRL}Modularized Reinforcement Learning}
\begin{figure}
		\centering
		\includegraphics[width=12cm]{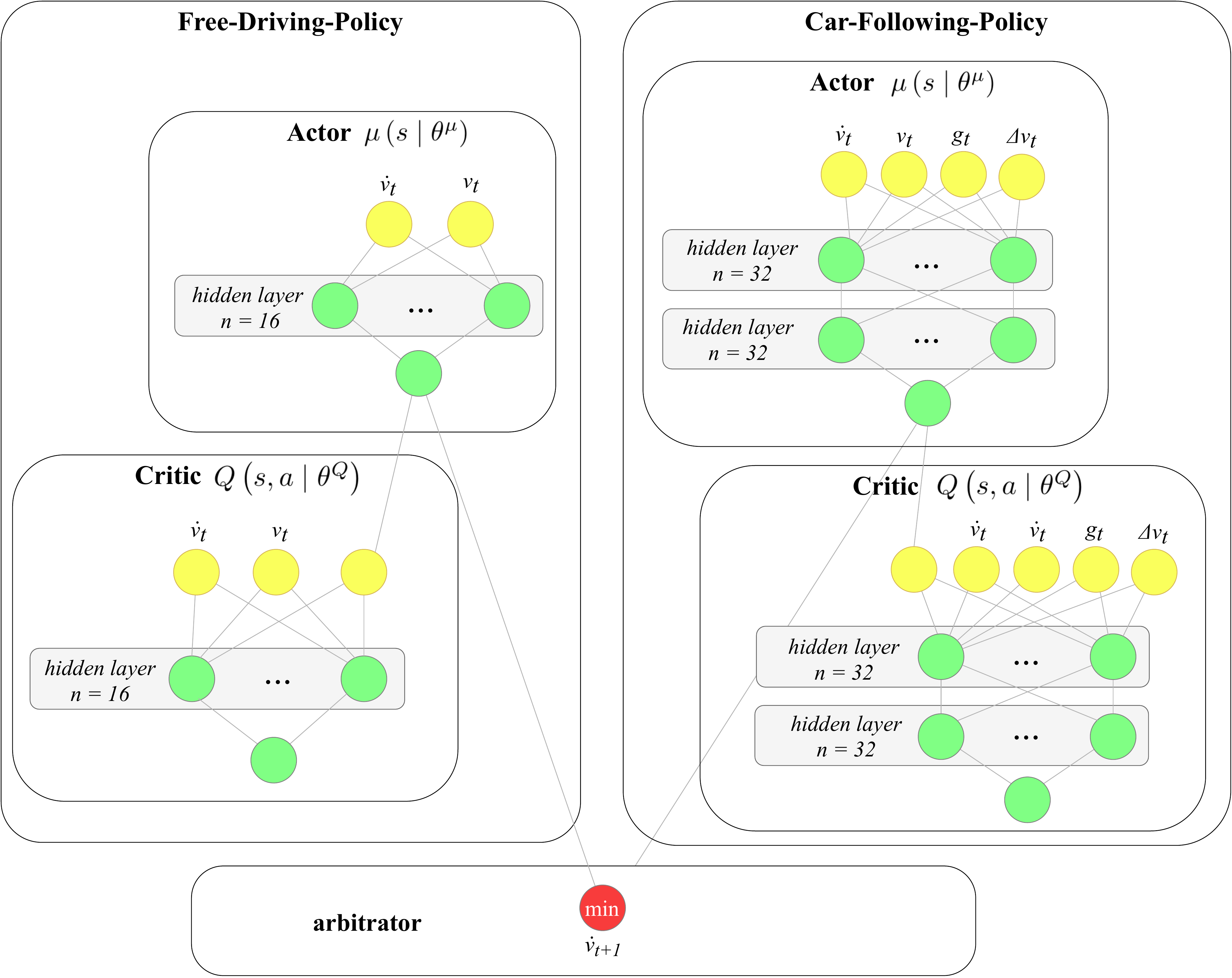}
		\caption{Modularized RL architecture with actor critic networks of both policies.} 
		\label{fig:MRL}
	\end{figure}

	Furthermore, we used a modularized approach to decompose the task into two subtasks. Modular Reinforcement Learning (MRL) refers to the decomposition of a complex, multi-goal
	problem into a collection of simultaneously running single goal learning processes, typically modeled as Markov Decision Processes. Typically, these subagents share an action set but have their own reward signal and state space. At each
	time step, every subagent reports a numerical preference for
	each available action to an arbitrator, which then selects one
	of the actions for the agent as a whole to take (\cite{MRL}). Numerous works are using a modularized Reinforcement Learning approach, like in \cite{MRLexample1}, \cite{MRLexample2} or \cite{MRLexample3} just to name a few. The advantage of decomposing multiple-goal reward functions with MRL, we also want to use in this work. 
	Figure \ref{fig:MRL} shows the architecture of our MRL System. We divide our car-following problem into two subtasks, handled by two different policies: 
	\begin{itemize}
		\item the Free-Driving Policy refers to free driving and aims to not to exceed a desired speed
		\item the Car-Following Policy refers to following a vehicle and aims to keep a reasonable gap to a leader vehicle  	
	\end{itemize}
	Although both policies are trained with different reward functions and in different training environments, they both output an acceleration value. As an arbitrator between both accelerations, we use a simple min-function. We adopted this approach from the IDM+ that also uses separate terms for free-flow and interaction with a leading vehicle, see \cite{idm_plus}. In the next sections, the model specifications of the Free-Driving Policy and the Car-Following Policy, both trained with the DDPG algorithm, are described in detail.
	
	\section{Free-Driving Policy}
	\label{sec:FreeDrivingPolicy}
	\subsection{Action and state space}
	\label{stateSpaceFree}
	The defined modularized RL approach requires that the action space of
	both sub-policies are identical. In our use case, the action space is
	continuous and one-dimensional and given by the range of feasible accelerations
	$\dot{v}_t \in [\dot{v}\sub{min}, \dot{v}\sub{max}]$. The range limits
	$\dot{v}\sub{min}= \unit[-9]{m/s^2}$ and  $\dot{v}\sub{max} =\unit[2]{m/s^2}$
	reflect comfortable driving ($\dot{v}_t \le \dot{v}\sub{max}$) while
	allowing for the physical limit $-\dot{v}\sub{min}$ of the braking
	deceleration in safety-critical
	situations.
	
	The state space defines the observations that the vehicle can receive from the environment. To compute an optimal
	acceleration, the following vehicle observes its own acceleration $\dot{v}_t$,
	and its own speed $v_t$ at time $t$. Linear translation and scaling are used to
	reduce the dimensions and to bring all observations approximately into
	the same range of $[0,1]$. The observation vector at time step $t$ is defined
	as 	
	\begin{equation}
		\label{eq:stateFree}
		s_t = \begin{pmatrix} \frac{v_t}{v\sub{des}} \\ \frac{\dot{v}_t-\dot{v}\sub{min}}{\dot{v}\sub{max} - \dot{v}\sub{min}}\end{pmatrix},
	\end{equation}
	where $v_{des}$ defines the desired speed.
	\subsection{Reward Function}
	\label{rewardFunctionFree}
	The reward function contribution at time step $t$ is decomposed into two factors. The first part aims to
	not exceed a desired speed $v\sub{des}$, but also to accelerate if the
	desired speed is not reached yet  
	\begin{equation}
		\label{eq:r1Free}
		r_{t, \rm speed}  = 
		\begin{cases}
			\frac{v_t}{v\sub{des}},
			& \text{if } v_t \leq v\sub{des}\\
			0,
			& \text{otherwise}.
		\end{cases}
	\end{equation}
	
	The second part of the reward function aims to reduce the jerk in
	order to achieve comfortable driving 
	\begin{equation}
		\label{eq:r2Free}
		r_{t, \rm jerk} = -\left(\frac{1}{j\sub{comf}} \ \dv{\dot{v}_t}{t}\right)^2,
	\end{equation}
	where $j\sub{comf}$ denotes a comfortable jerk of $\unit[2]{m/s^3}$.
	Since building up a comfortable value of acceleration or deceleration
	from zero in one second is at the limit of comfortable jerks, $r_{t, \rm jerk} $
	values above unity tend to feel uncomfortable.
	
	The contribution of the final reward function at simulation time step $t$ is the weighted
	sum of these factors according to
	\begin{equation}
		\label{rt1}
		r_{t, \rm free} =r_{t, \rm speed} + w\sub{jerk} r_{t, \rm jerk} ,
	\end{equation}
	where all the factors are evaluated at time step $t$.

	\subsection{Training environment}
	\label{training_environment1}
	In order to train the RL agent for the task of keeping a desired
	speed, the training environment is defined as follows. For the description of the vehicle dynamics, a point-mass kinematic model is used. The Euler and ballistic methods are used to update the speed and position for time step $t + 1$. This approach is recommended in \cite{numericalUpdateMethodsTreiber} as an efficient
	and robust scheme for integrating car-following models	
	\begin{align}
		v_{t+1} &= v_{t} + \dot{v}_{t} d, \\
		x_{t+1} &= x_{t} + \frac{v_{t} + v_{t+1}}{2} \Delta t,
	\end{align}
	with $\Delta t$ corresponding to the simulation step size that is globally
	set to \unit[100]{ms}. To train the RL agent, a training episode has to be defined. One training episode contains 500 time steps, and the vehicle's initial speed is set uniformly at random in the range $[0,v\sub{des}]$.
	
	\subsection{Model training}
	Both neural networks, critic and actor, are feed-forward neural
	networks with one layer of hidden neurons, containing 16 neurons (see
	Figure \ref{fig:MRL}). ReLU activation functions (\cite{relu}) are used, except for the output layer of the actor network that uses a tanh($\cdot$) activation function. The bounded output $a_t$ in the range $[-1,1]$ is then mapped to the acceleration range by
	\begin{equation}
		\dot{v}_t = \min \left( \abs{\dot{v}\sub{min}} a_t, \dot{v}\sub{max}\right).
	\end{equation}
	The learning rate for updating the weights of the critic and actor network is set to 0.001. For the exploration of the action space, an exploration noise model has to be defined. We adopted a
	zero-reverting Ornstein-Uhlenbeck process as suggested in \cite{DDPG}
	\begin{equation}
		\mathrm{d} x_{t}=-\theta_1 x_{t} \mathrm{d} t+\sigma_1 \mathrm{d} W_{t},
	\end{equation}
	with $\theta_1 = \unit[0.15]{s^{-1}}$, $\sigma_1 = \unit[0.2]{s^{-0.5}}$ and $W_{t}$ denoting the Wiener process.	
	The soft update rate of the target networks $\tau$ is set to
	0.001. All DDPG parameters are presented in Table \ref{tab:DDPGparameters}.
	\begin{table}
		\caption{DDPG parameter values} 
		\label{tab:DDPGparameters} 
		\begin{center}
			\begin{tabular}{ p{0.4\textwidth} p{0.2\textwidth}  p{0.2\textwidth} }
				& Free-Driving Policy & Car-Following Policy \\ \hline
				Learning rate & 0.001 & 0.001\\ 
				Reward discount factor & 0.95 & 0.95 \\ 
				Experience buffer length & 100000 & 100000 \\ 
				Mini batch size & 32 & 32 \\ 			
				Ornstein-Uhlenbeck  $\theta_1$ & 0.15& 0.15 \\ 
				Ornstein-Uhlenbeck  $\sigma_1$ & 0.2 & 0.2 \\				
				Number of hidden layers & 1 & 2\\
				Neurons per hidden layer & 16 & 32\\
				Soft target update  $\tau$ & 0.001 & 0.001\\

			\end{tabular}
		\end{center}
	\end{table}
	\begin{figure}
		\centering
		\includegraphics[width=12cm]{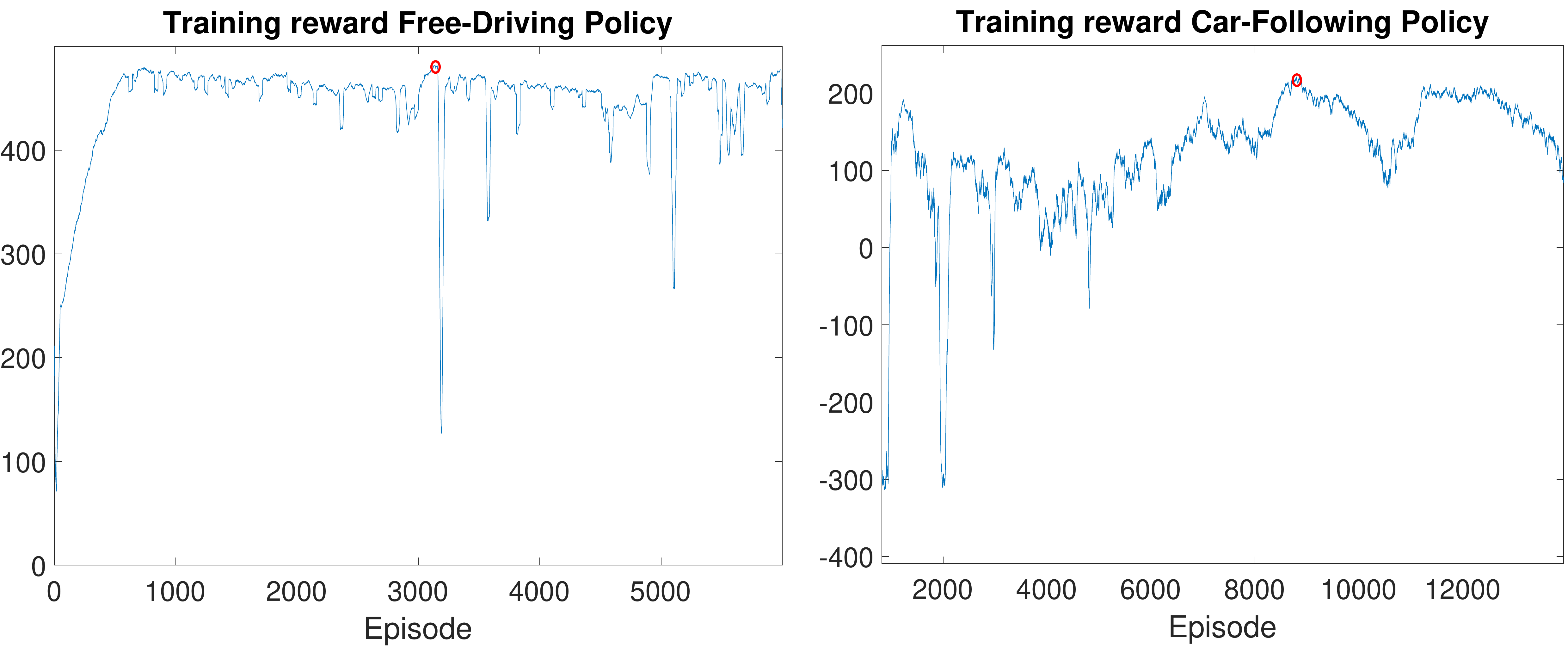}
		\caption{Asymmetric moving average reward of the last 30 training episodes for Free-Driving Policy and Car-Following Policy} 
		\label{fig:TrainingReward}
	\end{figure}
	
	Figure \ref{fig:TrainingReward} shows an example of the training
	process, where the asymmetric moving average reward of the last 30 training episodes is plotted over training episodes. For the Free-Driving Policy, a maximum average reward has been reached after approximately 3200 training episodes (marked in red).

	\section{Car-Following Policy}
	\label{sec:CarFollowingPolicy}
	\subsection{Action and state space}
	\label{stateSpaceFollow}
	To match the action space of the Free-Driving Policy, the acceleration is analogously defined as a continuous variable in the range between $\dot{v}\sub{min} = \unit[-9]{m/s^2}$ and $\dot{v}\sub{max} =\unit[2]{m/s^2}$.	
	
	The state space defines the observations that the following
	vehicle can receive from the environment. To compute an optimal
	acceleration, the following vehicle uses its own acceleration $\dot{v}_t$,
	its own speed $v_t$, the leader's speed $v_{t,l}$,
	and the (bumper-to-bumper) gap $g_t$.
	Again, linear translation and scaling are used
	to reduce the dimensions and to bring all observations approximately
	into the same range of $[-1,1]$. The observation at time step $t$ is defined as
	
	\begin{equation}
		\label{eq:stateFollow}
		s_t = 
		\begin{pmatrix} 
			\frac{v_t}{v\sub{des}} \\ 
			\frac{\dot{v}_t-\dot{v}\sub{min}}{\dot{v}\sub{max} - \dot{v}\sub{min}}\\
			\frac{v_{t,l}-v_t}{v\sub{des}}\\
			\frac{g_t}{g\sub{\rm max}}
			
		\end{pmatrix},
	\end{equation}
	where $g\sub{max}$ is set to  \unit[200]{m}. When $g_t$ exceeds  $g\sub{max}$ or there is no leader, $g$ is set to  $g\sub{max}$.
	
	\subsection{Reward Function}
	\label{rewardFunctionFollow}
	The goal of the Car-Follower-Policy is to reduce the crash risk while
	maintaining comfortable driving in non-safety-critical situations. The
	reward function includes a set of parameters that can be
	adjusted to realize different driving styles. 
	
	\begin{table}
		\caption{RL agent parameters and default values} 
		\label{tab:agentParameters} 
		\begin{center}
			\begin{tabular}{ p{0.12\textwidth}| p{0.65\textwidth}| p{0.1\textwidth}}
				Parameter & Description & Value \\ \hline
				$\dot{v}\sub{min}$ & Minimum acceleration & $\unit[-9]{m/s^2}$ \\  
				$\dot{v}\sub{max}$ & Maximum acceleration & $\unit[2]{m/s^2}$ \\  
				$b_{\rm comf}$ & Comfortable deceleration & $\unit[2]{m/s^2}$ \\  
				$j_{\rm comf}$ & Comfortable jerk & $\unit[2]{m/s^3}$ \\  
				$v_{\rm des}$ & Desired speed & $\unit[15]{ m/s}$ \\  		
				$T$ & Desired time gap to the leading vehicle & $\unit[1.5]{s}$ \\
				$g_{\rm min}$ & Desired minimum space gap & $\unit[2]{m}$ \\
				$T_{\rm lim}$ & Upper time gap limit for zero reward (see
				Equation~\eqref{eq:r1Free}) & $\unit[15]{s}$ \\
				$w\sub{gap}$ & relative weight for keeping the desired gap & 0.5\\
				$w\sub{jerk}$ & relative weight for comfortable acceleration & 0.004\\
			\end{tabular}
		\end{center}
	\end{table}

	The reward function contribution at time step $t$ consists of three factors. 
	The first factor addresses the driver's
	response in safety-critical situations by comparing the
	kinematically needed deceleration (assuming an
	unchanged speed of the leader) with the
	comfortable deceleration $b\sub{comf}$

	\begin{equation}
		\label{eq:r1_CFP}
		r_{t, \rm safe} = 
		\begin{cases}
			-\tanh\left(\frac{b\sub{kin}-b\sub{comf}}{-\dot{v}\sub{min}}\right),& \text{if } b\sub{kin}>b\sub{comf}\\
			0,              & \text{otherwise}
		\end{cases}
	\end{equation}
	
	with
	\begin{equation}
		\label{bkin}
		b\sub{kin} = 
		\begin{cases}
			\frac{(v_t-v_{t,l})^2}{g_t},& \text{if } v_t>v_{t,l}\\
			0,              & \text{otherwise},
		\end{cases}
	\end{equation}
	the kinematic deceleration $b\sub{kin}$ representing the minimum deceleration necessary to avoid a collision.
	The argument of the tanh($\cdot$) function with  the
	maximum possible deceleration ($\approx \unit[9]{m/s^2}$ on dry roads) gives a
	non-dimensional measure for the seriousness of the critical situation
	with values 
	near or above 1 indicating an imminent crash.  The tanh($\cdot$) function functions as a limitation for the reward value to the range $[-1,0]$. This has shown to make the learning process more stable. Notice that the case distinction in ~\eqref{eq:r1_CFP}  ensures that
	this term is not activated in non-critical situations. The purpose of
	the factor $r_{t, \rm safe}$ is twofold: It motivates the follower vehicle to
	brake in safety-critical or near-crash situations.  Furthermore, it motivates the follower vehicle also to break early in noncritical situations if the expected needed deceleration is above the comfortable, in order to achieve a comfortable
	approaching of the leader vehicle.
	
	The second part of the reward function aims to not fall below a reasonable
	distance to the leading vehicle

	\begin{equation}
		\label{eq:r2_CFP}
		r_{t, \rm gap}  = 
		\begin{cases}
			\frac{\varphi\left\{(g_t-g\sub{opt})/g\sub{var}\right\}}{\varphi(0)}, & \text{if } g_t < g^*\\
			\frac{\varphi\left\{(g_t-g\sub{opt})/g\sub{var}\right\}}{\varphi(0)}\left(1-\frac{g_t-g^*}{g\sub{lim} - g^*}\right)  & \text{otherwise},
		\end{cases}
	\end{equation}
	with 
	\begin{align}
		g\sub{opt} &= v_tT + g\sub{min},\\
		g\sub{var} &= 0.5g\sub{opt},\\
		g\sub{lim} &= v_tT\sub{lim} + 2g\sub{min},
	\end{align}
	and $\varphi(x)$ describing the density function of the standard normal distribution. The value of $g^*$ is chosen in a way that the reward function $r_{t,\rm gap}$ is differentiable.
	\begin{figure}
		\centering
		\includegraphics[width=12cm]{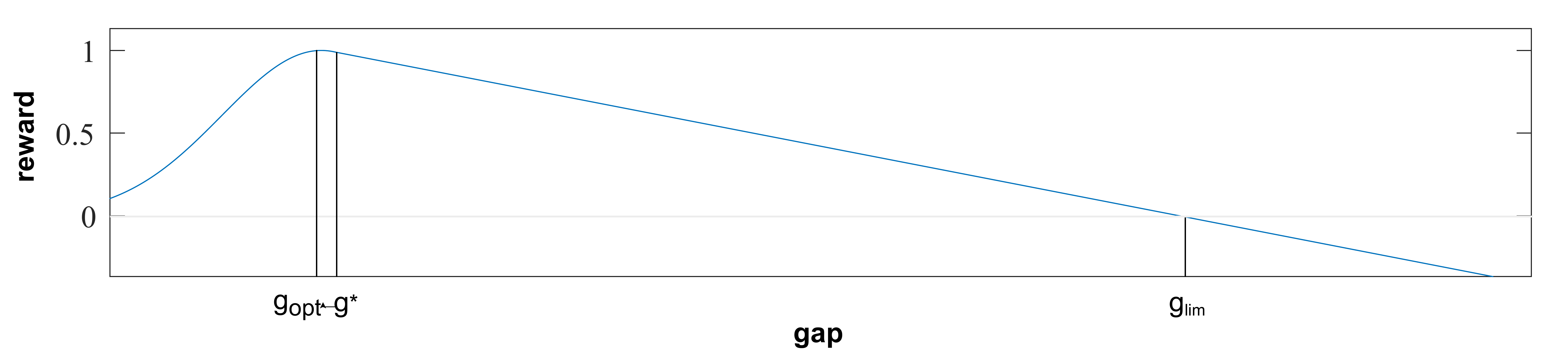}
		\caption{Factor $r_{t, \rm gap}$ of the reward function maximizing the reward
			for car following with time gap $T$} 
		\label{fig:RewardFunc1}
	\end{figure}
	Figure \ref{fig:RewardFunc1} illustrates the reward function for
	$r_{t,\rm gap}$, containing the parameter $g\sub{opt}$, $g^*$ and $g_{\rm
		lim}$. The reward function is designed in a way that for high speeds $v$
	of the following vehicle, the time gap between following and leading
	vehicle tends to $T$, while for low speeds, the distance
	between both tends to $g\sub{min}$. Different values of $T$
	result in different driving styles in a way that, for lower values of
	$T$ and $g\sub{min}$, the driver follows
	more closely the leading vehicle resulting in a more aggressive
	driving style. The results for different values of $T$ can
	be found in Section \ref{sec:differentT}. Different functions for $g_t
	> g^*$ have been applied, but the best results regarding a smooth and
	comfortable approaching of the following vehicle have been reached with
	a linear function. Furthermore, a high value of $T\sub{lim}$ has been
	chosen, resulting in a low gradient of the linear function. This
	prevents the follower vehicle from trying to reach the desired time gap $T$
	as fast as possible with maximum speed but rather motivates the follower vehicle to decelerate early and approach the leading vehicle comfortably.

	The third factor of the reward function aims to reduce the jerk in
	order to achieve comfortable driving and has been designed analogously to the Free-Driving Policy
	\begin{equation}
		\label{eq:r3_CFP}
		r_{t, \rm jerk} = -\left(\frac{1}{j\sub{comf}} \ \dv{\dot{v}_t}{t}\right)^2.
	\end{equation}
	The contribution of the final reward function~\eqref{Rt}  at simulation time step $t$ is the weighted
	sum of these factors according to
	\begin{equation}
		\label{rt2}
		r_{t, \rm follow} = r_{t, \rm safe} + w\sub{gap}r_{t,\rm gap}+w\sub{jerk}r_{t, \rm jerk} ,
	\end{equation}
	where all the factors are evaluated at time step $t$. The weights (cf.
	Table~\ref{tab:agentParameters}) have been found experimentally and
	will be optimized in future studies.

	\subsection{Training environment}
	\label{training_environment2}
	The training environment is modeled by a leader and a follower vehicle. Both vehicles implement the point-mass kinematic model described in Section~\ref{training_environment1}. While the RL agent controls the follower vehicle, the leading trajectory is based on an Ornstein–Uhlenbeck process, whose parameters
	reflect the kinematics of real leaders. The Ornstein–Uhlenbeck process describes
	the speed of the leading vehicle $v_{t,l}$ and is defined as 	
	\begin{equation} \label{eq:AR1}
		\mathrm{d} v_{t,l}=\theta_2\left(\mu_2-v_{t,l}\right) \mathrm{d} t+\sigma_2 \mathrm{d} W_{t},
	\end{equation}
	with $\theta_2 = \unit[0.132]{s^{-1}}$, $\mu_2 = \unit[7.5]{m/s}$, $\sigma_2 = \unit[3.847]{m/s^{1.5}}$ and $W_{t}$ denoting the Wiener process. The parameter have been designed to model the behavior of a real leader vehicle.
	For the numerical simulation, the Ornstein-Uhlenbeck process has been discretized using the Euler-Maruyama method \citep{CIRprocess}. The resulting discretized process is defined as
	\begin{equation} 
		v_{l}(n+1)=v_{l}(n)+\theta_2\left\{\mu_2-v_{l}(n)\right\} \Delta t+\sigma_2 \Delta W_{n},
	\end{equation}
	with 
	\begin{equation}
		\Delta W_{n} \sim N(0, \Delta t).
	\end{equation}

\begin{figure}
		\centering
		\includegraphics[width=0.95\textwidth]{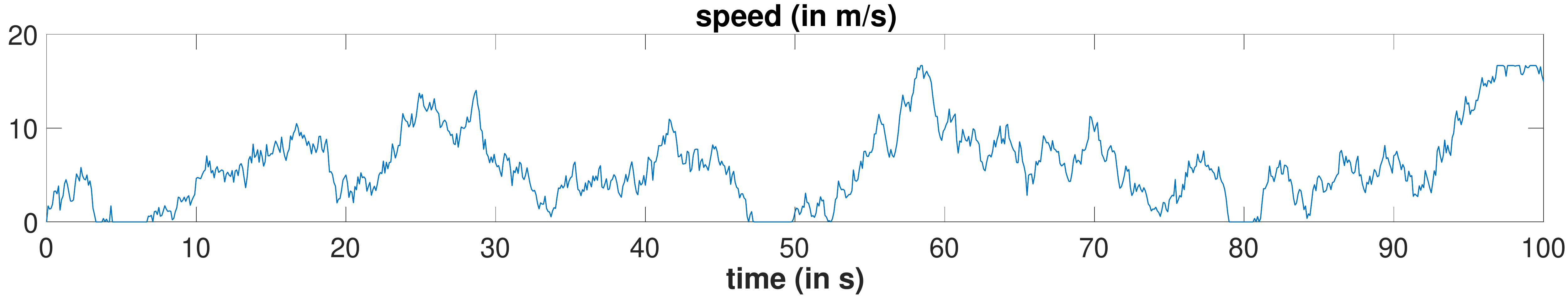}
		\caption{Example of a leading trajectory based on the parametrized Ornstein-Uhlenbeck process used to train the RL agent}
		\label{fig:AR1process}
	\end{figure}

	Figure \ref{fig:AR1process} shows an example trajectory of the leading
	vehicle. After the Ornstein-Uhlenbeck process is calculated for one
	episode, all speed values are clipped to the range $[0,\unit[16.6]{m/s}]$. This generates
	training situations, where the leader vehicle stands still for some
	time, e.g. at $t= [\unit[48]{s}, \unit[50]{s}]$ in Figure
	\ref{fig:AR1process}. Furthermore, the leader's trajectory
	also contains intervals where the speed is above the desired speed
	$v\sub{des}=\unit[15]{m/s}$ of both the leader and the follower,
	e.g., around $t=\unit[58]{s}$ and $t\in [\unit[97]{s},
	\unit[100]{s}]$.
	One training episode has a simulation time of \unit[50]{s}.  With a step size of
	$\Delta t=\unit[100]{ms}$, this results in an episode length of $500$
	steps. The initial speeds of the following and leading vehicles are both
	set uniformly at random in the range $[0,v\sub{des}]$,
	respectively. The initial space gap $g_0$ between both is set to \unit[120]{m}. 
	
	\subsection{Model training}
	Both neural networks, critic and actor, are feed-forward neural networks with two layers of hidden neurons, containing 32 neurons each (see Figure \ref{fig:MRL}). Except for this configuration, the architecture of the critic and actor, as well as the parameter of the DDPG algorithm, is the same as we used for the Free-Driving Policy (see Table \ref{tab:DDPGparameters}).
	
	Figure \ref{fig:TrainingReward} shows an example of the training
	process, where the asymmetric moving average reward of the last 30
	training episodes are plotted over training episodes. For the
	Car-Following Policy, a maximum average reward has been reached after
	approximately 8900 training episodes (marked in red). The training processes for both policies are quite unstable. As this is a known issue using the DDPG algorithm, we plan to use a more stable algorithm like the TD3 algorithm \citep{TD3}.

	\section{Validation}
	\label{sec:validation}
	The goal is not to minimize some error measure as in usual calibration\slash validation but to check if the driving style is safe,
	effective, and comfortable. The RL strategy is evaluated with respect to these metrics in different driving scenarios, described in the following.
	
	\subsection{Response to an external leading vehicle speed profile}
\begin{figure}
		\centering
		\includegraphics[width=0.95\textwidth]{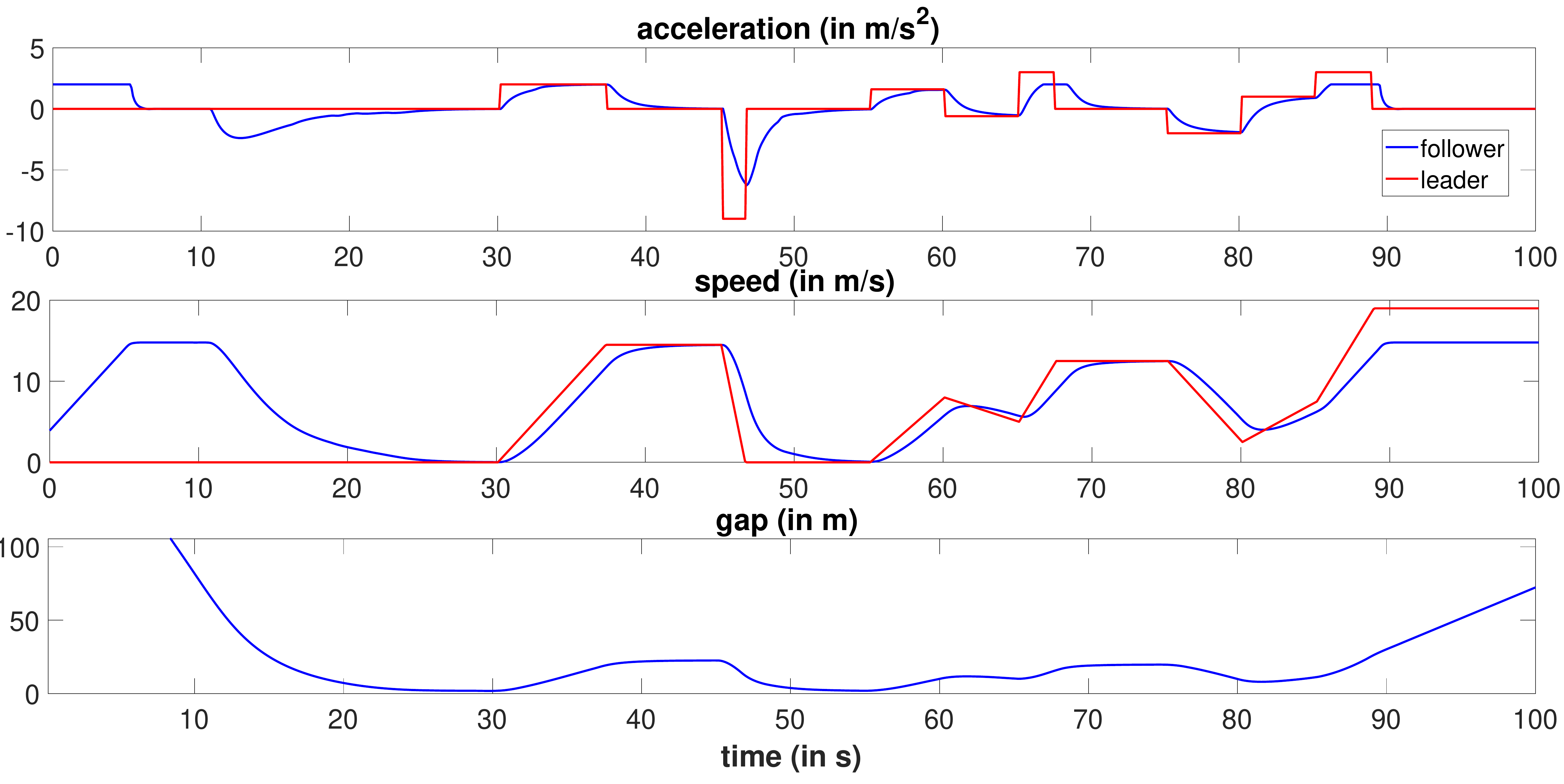}
		\caption{Response to an external leading vehicle speed
			profile.}
		\label{fig:manipulatedLeader}
	\end{figure}

	The first scenario is designed to evaluate the transition between free driving and car-following as well as the follower's behavior in safety-critical situations. 
	Figure \ref{fig:manipulatedLeader} shows a driving scenario with an
	artificial external profile for the leading vehicle speed. The initial
	gap between 
	follower and leader is 200 meters referring to a free driving
	scenario. The follower accelerates with $\dot{v}\sub{max} = \unit[2]{m/s^2}$ until
	the desired speed $v\sub{des} = \unit[15]{m/s} $ is reached and approaches
	the standing leading vehicle. When the gap between both drops below 
	\unit[70]{m},
	the follower starts to decelerate with a maximum
	deceleration of approximately $b_{\rm
		comf} = \unit[2]{m/s^2}$ (transition between free driving and car-following)
	and comes to a standstill with a final gap of approximately 
	$g\sub{min} = \unit[2]{m}$. Afterwards ($t=\unit[30]{s}$),  the leading vehicle accelerates to a speed
	that is below the desired speed of the follower before performing a
	maximum braking maneuver ($\dot{v}_l=\unit[-9]{m/s^2}$) to a full stop ($t=\unit[46]{s}$). At the time of the start of the
	emergency braking, the follower has nearly reached a steady
	following mode at the desired space gap $g_t=g\sub{min}+v_{t} T$. While this
	gap makes it impossible to keep the deceleration in the comfortable
	range without a rear-end collision, the follower makes the best of
	the situation by braking smoothly with a maximum deceleration of $-\dot{v}
	\approx \unit[5]{m/s^2}$.  The transition between different
	accelerations happens in a comfortable way reducing the resulting
	jerk. Only at the beginning ($t=\unit[46]{s}$) where the situation is
	really critical, the jerk $\text{d}\dot{v}/\text{d}t$ exceeds the comfortable range 
	$\pm \unit[1.5]{m/s^3}$. Afterwards, the leader performs a series of
	non-critical acceleration and deceleration maneuvers and the follower
	tries to follow the leader at the speed dependent desired space gap
	$g\sub{min}+v_{t}T$ while simultaneously smoothing the leader's speed profile. After the leaders speed exceeds the desired speed of the follower at $t=\unit[88]{s}$ (transition between car-following and free driving), the follower keeps the desired speed $v\sub{des} = \unit[15]{m/s} $.

	\begin{figure}
		\centering
		\includegraphics[width=0.95\textwidth]{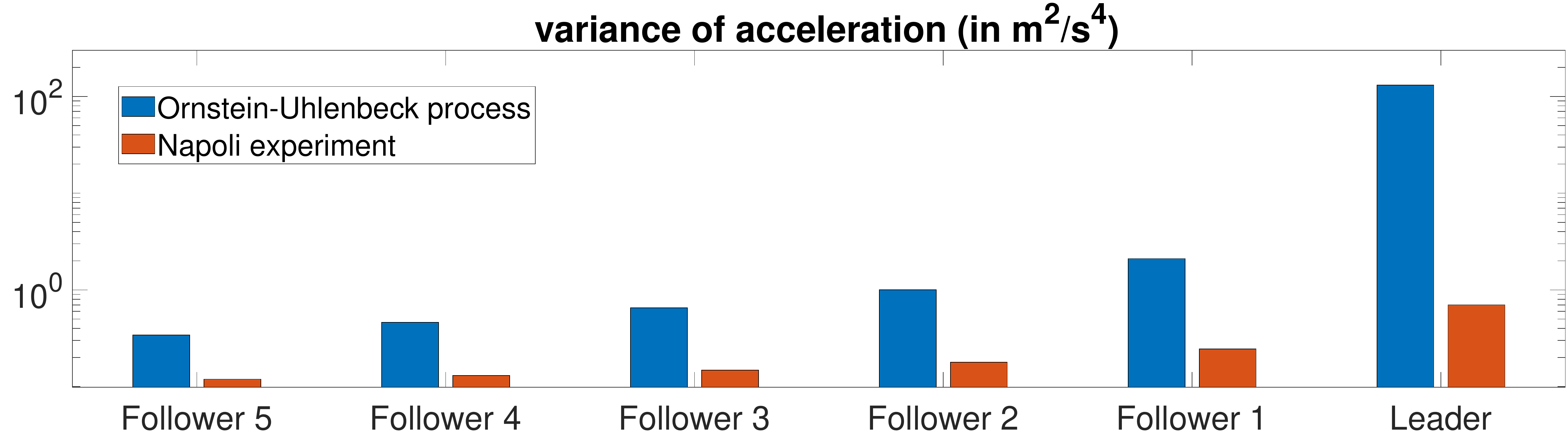}
		\caption{Comparison of the acceleration variance between
			leader and follower for a leader controlled by the Ornstein-Uhlenbeck process (blue
			bars) and the leading vehicle of the Napoli experiment
			(red).}
		\label{fig:VarAccComp}
	\end{figure}

	\subsection{String stability}
	\label{sec:stringStability}
\begin{figure}
		\centering
		\includegraphics[width=0.95\textwidth]{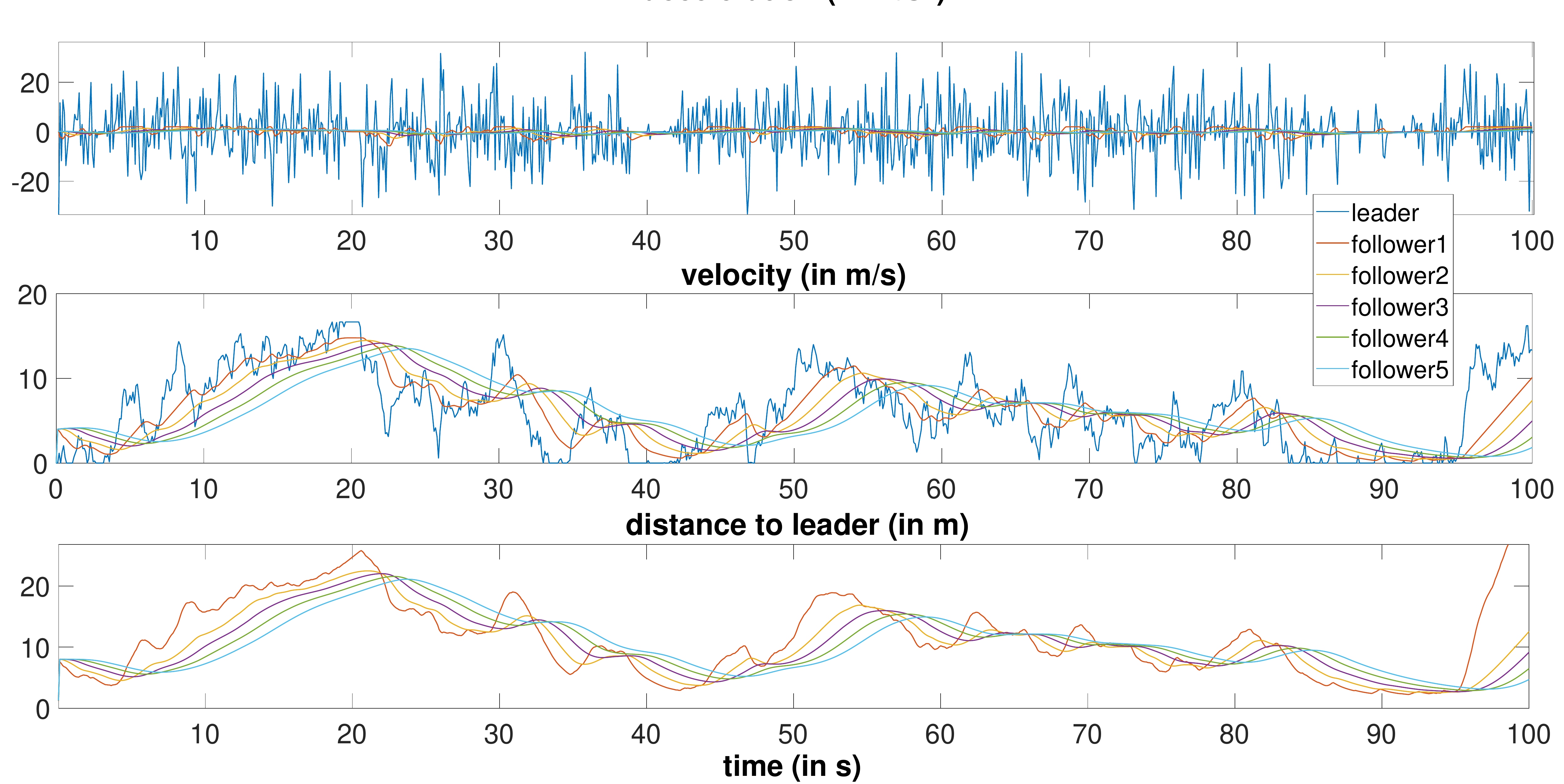}
		\caption{Response to a leader trajectory based on a Ornstein-Uhlenbeck process}
		\label{fig:AR1Kolonne}
	\end{figure}

	The second validation scenario, shown in Figure
	\ref{fig:AR1Kolonne}, consists of a leader based on the Ornstein-Uhlenbeck process from Section \ref{training_environment2}
	that is
	followed by five vehicles, each controlled by the trained RL
	agent. The results show that traffic oscillations can effectively be
	dampened with a sequence of trained followers, even if the leader
	shows large outliers in acceleration. Figure \ref{fig:VarAccComp}
	illustrates the difference of accelerations between the leader and the
	followers (blue bars). The last follower shows the lowest variance of
	acceleration demonstrating the ability of the RL agent to
	flatten the speed profile, to dampen oscillations, and thus to increase
	comfort and reduce fuel consumption and emissions.

	\subsection{Response to a real leader trajectory}
	
\begin{figure}
		\centering
		\includegraphics[width=0.95\textwidth]{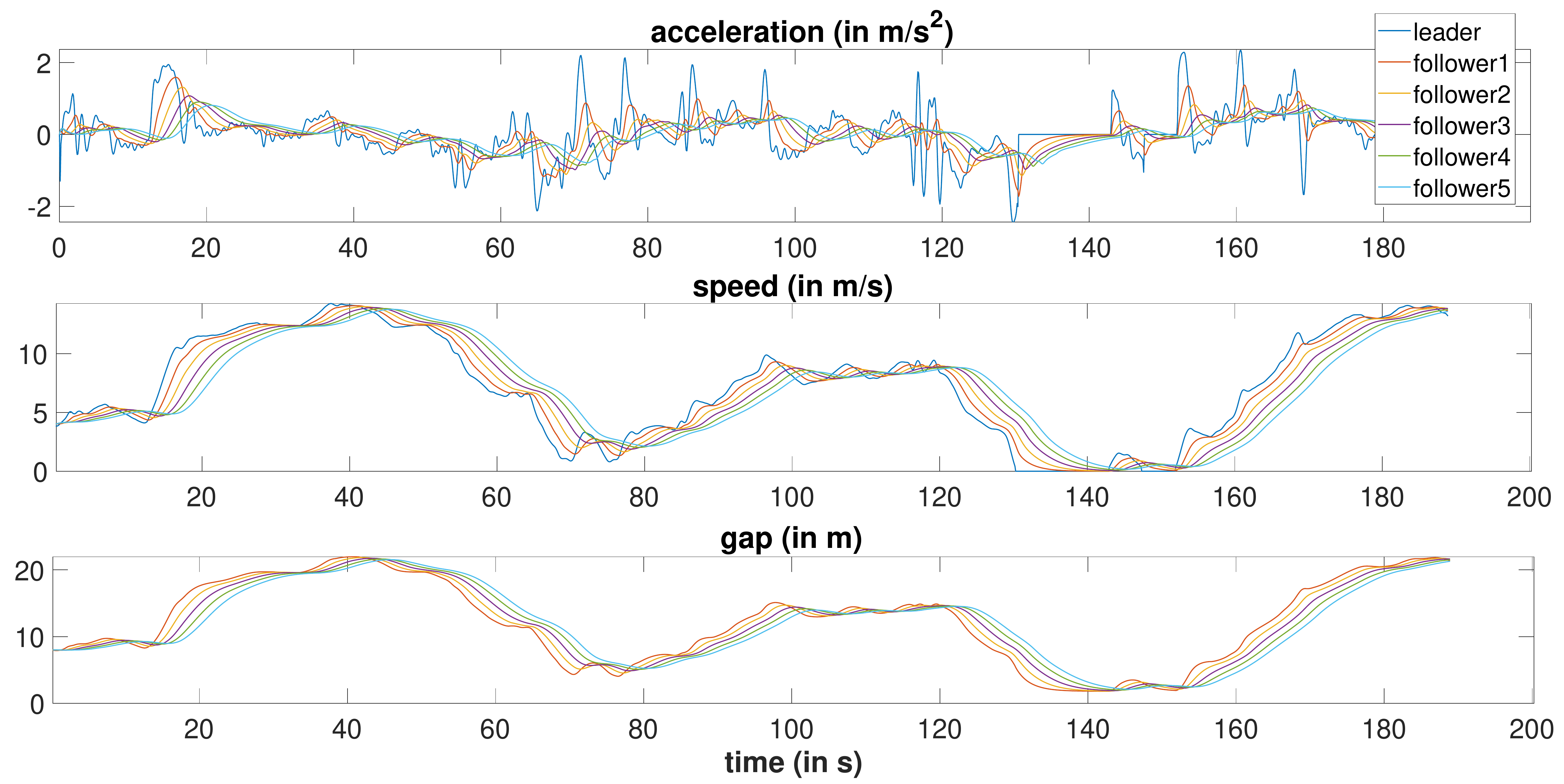}
		\caption{Response to a real leader trajectory}
		\label{fig:PunzoKolonne}
	\end{figure}

	In a further scenario, the abilities of the RL strategy are evaluated
	with a real leader trajectory (Figure~\ref{fig:PunzoKolonne}). This
	trajectory comes from the platoon driving experiments of
	\cite{punzo2005nonstationary} on urban and peripheral arterials in Napoli
	where high-precision distance data between the platoon
	vehicles were obtained. Similar to the experiment from
	Section~\ref{sec:stringStability}, string stability and the reduction
	of  the acceleration variance, shown by the red bars in Figure
	\ref{fig:VarAccComp}, is demonstrated. At time $t = \unit[140]{s}$ the leader
	stands still, and it can be observed that all following vehicles are
	keeping the minimum distance $g\sub{min}$ to the leader.  
	
	Comparing the first three RL followers with the three followers of the real experiment, we notice that the RL followers drove more comfortably with less acceleration than the real drivers. This confirms our approach to not calibrate existing trajectories  directly and maximizing an external reward function instead

	\subsection{Response of different driver characteristics}
	\label{sec:differentT}
	
\begin{figure}
		\centering
		\includegraphics[width=0.95\textwidth]{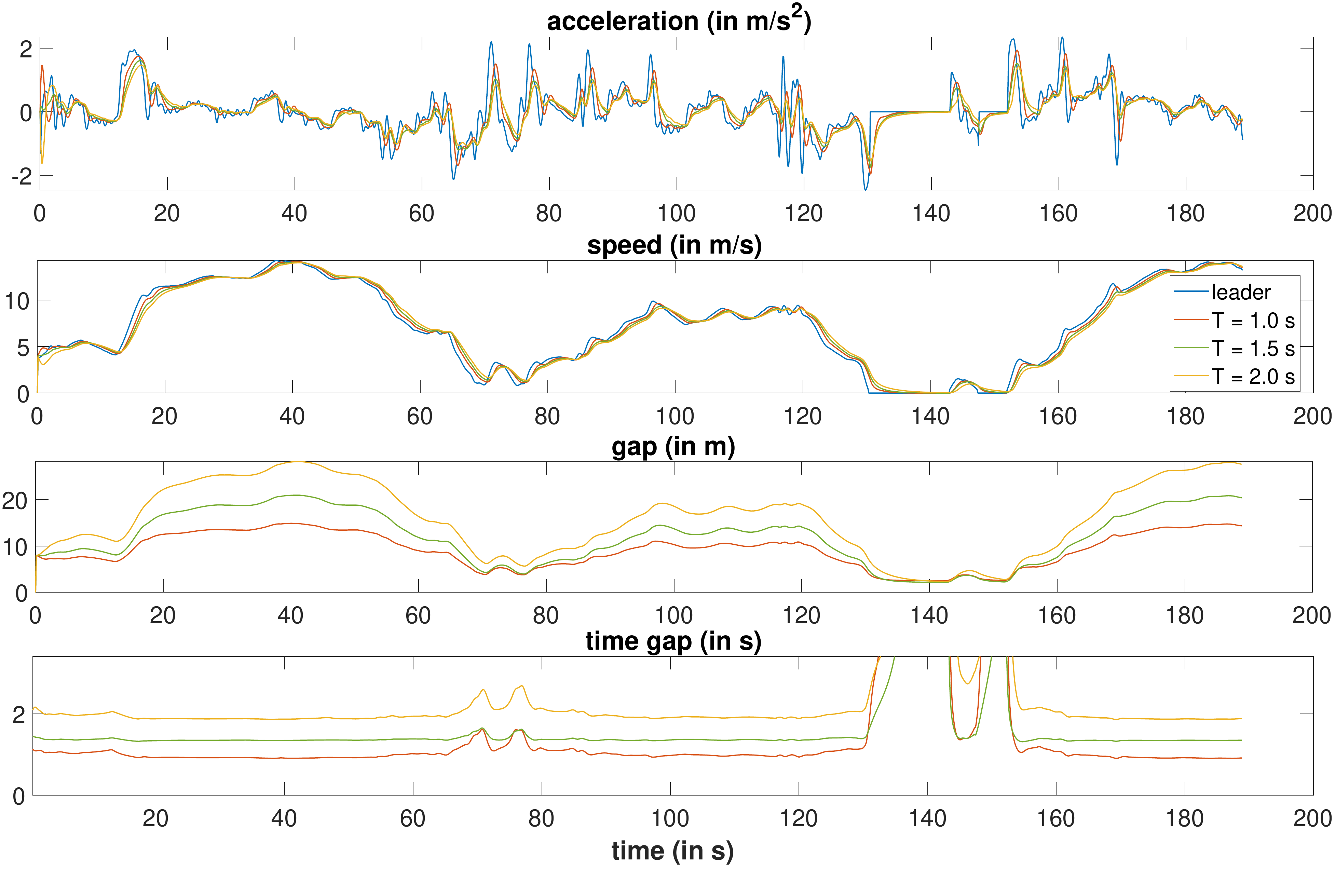}
		\caption{Impact of differently parametrized RL agents
			on the driving behavior }
		\label{fig:differentT}
	\end{figure}

	As mentioned in Section \ref{rewardFunctionFollow}, different driving styles
	can be achieved by adjusting the parameters of the reward
	function. Three RL agents have been trained on a reward function, that
	differs in the desired time gap $T$ between following and
	leading vehicle ($T_{1} = 1.0s$, $T_{2} = 1.5s$, 
	$T_{3} =2.0s$). Figure \ref{fig:differentT} shows the result of these agents,
	following the real leader trajectory from Napoli. It can be observed,
	that a lower value for $T$ results in closer driving to the
	leader. The last plot in Figure \ref{fig:differentT} which illustrates the actual time gap $ (g_t-g\sub{min})/v_t$ shows, that for speeds greater than zero the desired time gap $T$ is approximately maintained.
	This behavior can be considered as a more "aggressive"
	driving style. Since this also means that there are fewer options in
	increasing driving comfort without affecting safety, the follower's
	accelerations and decelerations are higher than that of the more
	conservative RL agent~3 although the relative
	weighting $w\sub{jerk}/w\sub{gap}$ of the safety and comfort aspects in Equation~\eqref{rt1} and \eqref{rt2} has not been
	changed. Still, when simulating platoons of the three RL
	agents in any order, the accelerations and jerks decrease along the
	platoon.
	
	\subsection{Cross validation with the Intelligent Driver Model}
	\label{sec:crossValIDM}
	To compare the performance of the RL agent with that of
	classical car-following models, we chose the commonly used
	IDM of \cite{Opus} whose acceleration is given
	by 
	\begin{equation}
		\label{eq:IDM}
		\dv{v}{t}=\dot{v}\sub{max}\left[1-\left(\frac{v_t}{v\sub{des}}\right)^{4}-\left\{\frac{s^{*}\left(v_t, v_{t,l}\right)}{g_t}\right\}^{2}\right],
	\end{equation}
	with
	\begin{equation}
		\label{eq:IDMsstar}
		s^{*}\left(v_t, v_{t,l}\right)=g\sub{min}+\max \left\{0,v_tT+\frac{v_t(v_t-v_{t,l})}{2 \sqrt{\dot{v}\sub{max} b\sub{comf}}}\right\}.
	\end{equation}
	Notice that the IDM parameters desired
	speed $v\sub{des}$, minimum gap $g\sub{min}$, time gap $T$, maximum
	acceleration $\dot{v}\sub{max}$, and
	comfortable deceleration $b\sub{comf}$ are a subset of that of the RL reward
	function. 
	
	First, we calibrate the IDM on the Napoli data set by
	minimizing the sum of squares of the relative gap error,
	$\mathrm{SSE}(\ln g)$, of the first follower with respect to the
	data (cf. Table~\ref{tab:IDMparameters}). The same parameters are also
	assumed for the reward function of the RL agent before it was trained
	on the artificial Ornstein-Uhlenbeck generated leader speed profile. Notice that the RL agent used the Napoli data only
	indirectly by parameterizing its reward function.
	
	\begin{table}
		\caption{IDM parameters calibrated to the Napoli
			data and also used for the reward function of the RL agent}
		\label{tab:IDMparameters} 
		\begin{center}
			\begin{tabular}{ p{0.14\textwidth} |p{0.1\textwidth}  } 
				Parameter & Value   \\ \hline
				$T$ & $\unit[0.83]{s}$\\
				$g\sub{min}$ & $\unit[4.90]{m}$\\
				$\dot{v}\sub{max}$ & $\unit[4.32]{m/s^2}$\\
				$b\sub{comf}$ & $\unit[2.34]{ m/s^2}$\\
				$v\sub{des}$ & $\unit[33.73]{m/s}$			
			\end{tabular}
		\end{center}
	\end{table}
	
\begin{figure}
		
		\centering
		\includegraphics[width=0.95\textwidth]{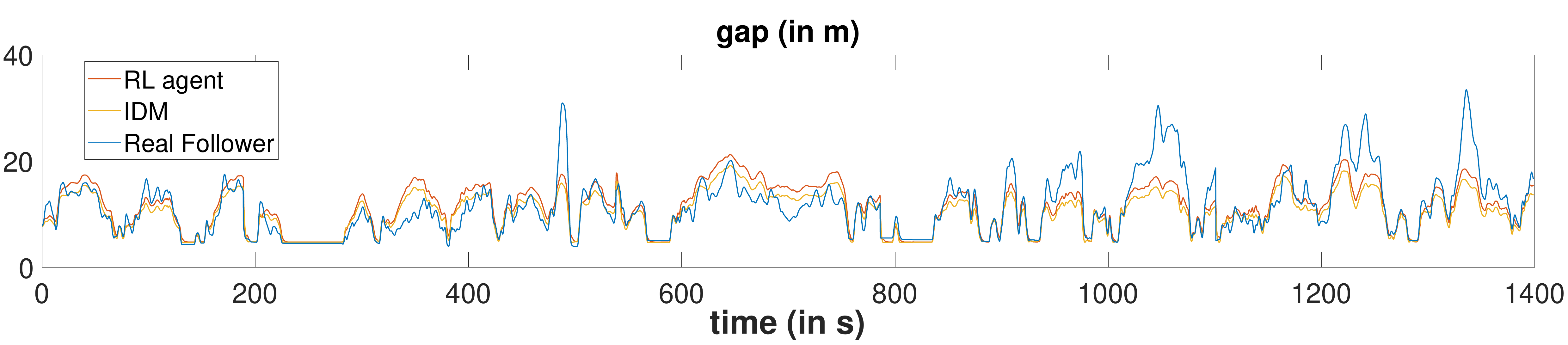}
		\caption{Comparison between IDM and RL agent, calibrated on the same set of parameters}
		\label{fig:IDMvsRL}
	\end{figure}

	Figure \ref{fig:IDMvsRL} shows the results for (i) the RL
	agent, calibrated on the real follower data (red lines), (ii) the IDM,
	calibrated on the same follower data (amber), and (iii) the real
	follower of the Napoli experiment (red). To compare the
	performance for both approaches, the respective objective functions
	have been computed. The objective function of the RL agent corresponds
	to the reward function, while the goodness-of-fit function
	$\mathrm{SSE}(\ln g)$ defines the objective function of the
	IDM. Furthermore, we cross-compared the values by calculating the
	reward function for the IDM and $\mathrm{SSE}(\ln g)$ for the RL model. All
	values are shown in Table \ref{tab:objectiveFunc}. 
	As expected, the RL agent performs better than the IDM relative to the
	reward function used for its learning. It is remarkable, however, 
	that the RL model also outperforms the IDM relative to the
	the goodness-of-fit function that has been used only indirectly by
	parameterizing its reward function.
	
	\begin{table}
		\caption{Comparison between calibrated RL agent and IDM for
			accumulated Reward and Goodness-of-Fit Function $\mathrm{SSE}(\ln g)$} 
		\label{tab:objectiveFunc} 
		\begin{center}
			\begin{tabular}{p{0.3\textwidth} | p{0.2\textwidth} p{0.2\textwidth}  } 
				& RL agent & IDM   \\ \hline
				$\mathrm{SSE}(\ln g)$ & $389.10$ &  $418.05$	\\
				Accumulated Reward &  $6.86 \times 10^3$   & $6.73\times 10^3$			
			\end{tabular}
		\end{center}
	\end{table}
	
	\subsection{Time-to-collision comparison with the IDM}
	
\begin{figure}	
		\centering
		\includegraphics[width=0.95\textwidth]{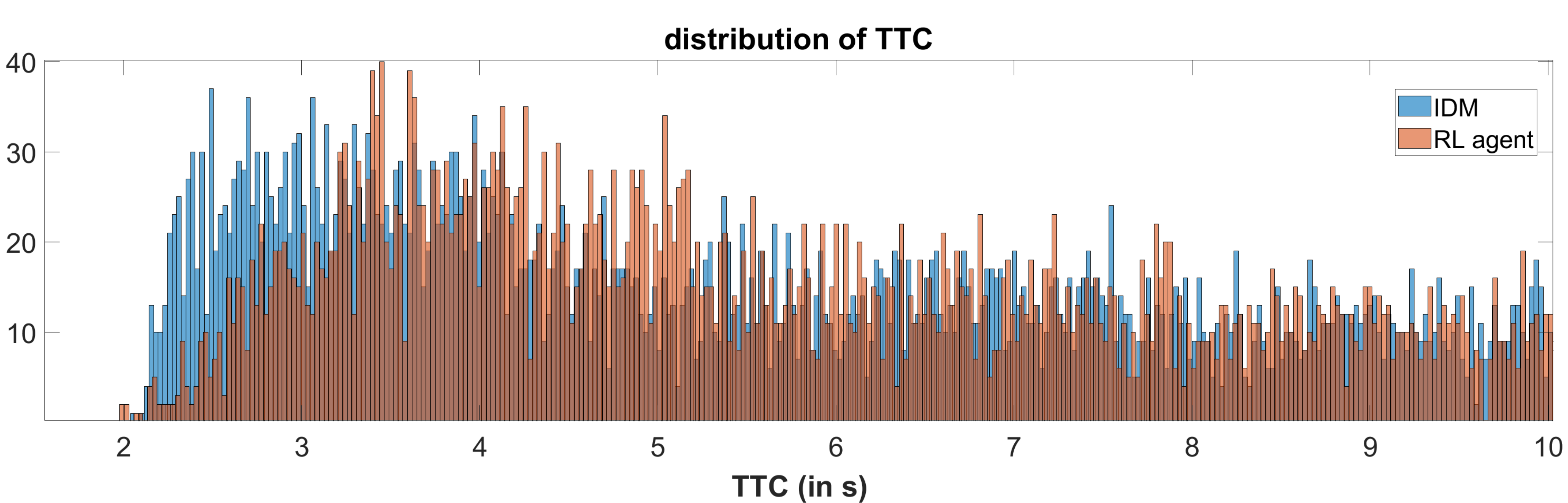}
		\caption{Distribution of time-to-collision (TTC) for IDM and RL agent in different experiments based on real leader trajectories}
		\label{fig:DistributionTTC}
	\end{figure}

	An important safety measure for car-following models is the
	time-to-collision (TTC). Again we took leading trajectories coming
	from platoon driving experiments in Napoli to conduct a TTC
	analysis. We simulated 15 different leaders from those experiments,
	once followed by an RL agent and the other time by an IDM vehicle (Sect. \ref{sec:crossValIDM}). Both
	followers are parametrized according to the default parameters of Table
	\ref{tab:agentParameters}. Figure \ref{fig:DistributionTTC} shows the
	distribution of TTC for all 15 experiments. The TTC has been evaluated
	at each simulation time step. The lower TTC bound for the RL agent is
	\unit[1.99]{s}, while the one for the IDM vehicle is
	\unit[2.04]{s}. But except for some values around TTC = \unit[2]{s}, the TTC values of the RL agent show to be higher than those of the IDM.
	
\begin{figure}	
		\centering
		\includegraphics[width=0.95\textwidth]{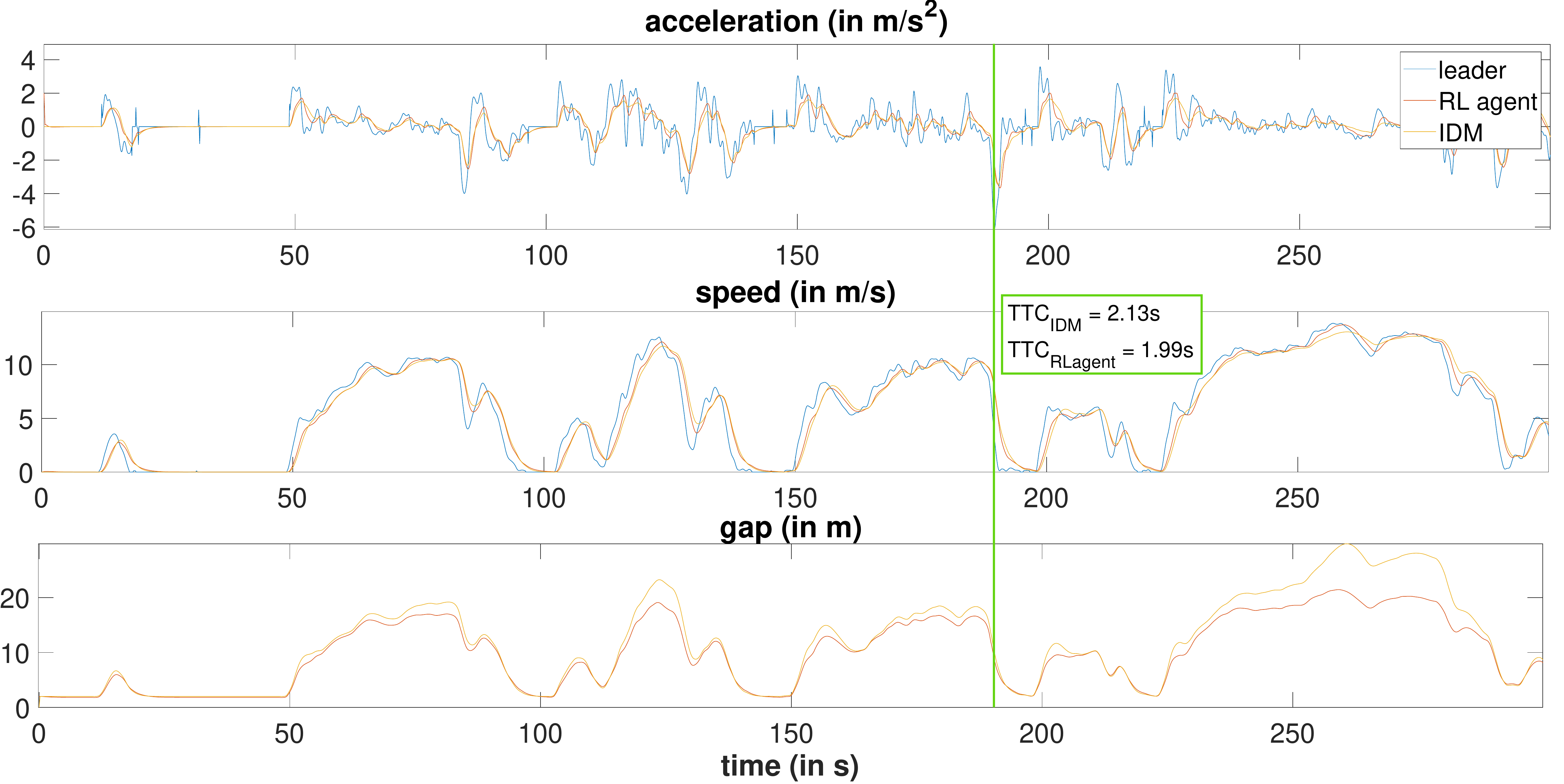}
		\caption{Response of an IDM and RL agent on a real trajectory. Safety-critical situations where the TTC of the IDM drops below \unit[2.5]{s} are marked in green.}
		\label{fig:TTC_CaseStudy}
	\end{figure}

	The difference in driving behavior is further
	evaluated by focusing on a single leader-follower experiment, shown in
	Figure \ref{fig:TTC_CaseStudy}, where the situation with the lowest TTC for the RL agent is marked green. In this situation, the leader brakes to a standstill with a 
	quite high deceleration of $-\dot{v}_l \approx \unit[6]{m/s^2}$. The reason for the slight
	difference in TTC between IDM and RL agent lies in the fact that the
	IDM keeps a slightly higher gap to the leader most of the time,
	which simply reflects a more defensive driving style. 
	
	\section{Conclusion/Discussion}
	\label{sec:conclusion}
	This study presented a novel car-following model based on
	Deep Reinforcement Learning. The
	proposed model considers free driving, car-following, as well as the
	transition between both in a way that approaching the leading vehicle
	is smooth and comfortable. We used an approach called Modularized
	Reinforcement Learning (\cite{MRL}) to decompose the multi-goal
	problem into two subtasks. Two different RL policies have been
	designed using the Deep Deterministic Policy Gradient algorithm. The
	Free-Driving Policy aims to reach and not exceed a certain
	desired speed. The Car-Following Policy aims to keep a reasonable gap
	to a leader vehicle and keep the traffic situation safe. 
	
	For each policy, we defined separate reward functions reflecting traffic safety and comfort aspects. 
	Different driver characteristics can be modeled by adjusting the parameter of the reward function.	
	The proposed model is trained on leading trajectories based on an
	Ornstein-Uhlenbeck process. This leads to high generalization capabilities and model usage in a wider variety of traffic scenarios. Furthermore, the
	supply of learning data is unlimited.
	For the evaluation of the trained agents, different traffic scenarios
	based on both synthetic and
	real trajectory data have been simulated, including situations that
	bring the model to its limits. 
	In all cases, the car-following model proved to be accident-free and
	comfortable. Further scenarios showed that traffic oscillations could
	effectively be dampened with a sequence of trained followers, even if
	the leader shows large outliers in acceleration.
	A reason for this favorable behavior is that, unlike
	classical car-following models, RL-based models
	even learn responses that are not explicitly contained in their
	specification. For example, the only reward component depending on
	the relative speed is the safety component~\eqref{eq:r1_CFP}
	which only kicks in for large approaching rates. Still, the
	actual model also responds to small values of the relative
	speed since such behavior may prevent reaching a
	safety-critical situation within the prediction horizon.
	
	Besides driving comfort, string stability, and safety,
	the efficiency of the resulting traffic flow is important,
	particularly, the maximum flow through a
	bottleneck (maximum of the respective fundamental diagram) that can be 
	sustained and the outflow from a region of congested
	traffic. Furthermore, we have idealized the state dynamics by
	assuming that a vehicle can instantaneously take on the acceleration
	prescribed by the actor while the real control path is more
	complicated and non-negligible response times happen, particularly
	for positive accelerations. We expect that RL techniques play out
	their strengths in such more complex state dynamics. All this  will
	be investigated in a forthcoming paper.

	\subsection*{Acknowledgements}
	We thank Vincenzo Punzo for making available to us the experimental
	car-following data used in this paper. This work was partially funded by BAW - Bundesanstalt für Wasserbau (Mikrosimulation des Schiffsverkehrs auf dem Niederrhein)

	\bibliography{RL_vehicles_references}
	
\end{document}